# Geometric Monomial (GEM): a family of rational $2N$-differentiable activation functions

Eylon E. Krause[1,2]

eylonkr@colman.ac.il

Weizmann Institute of Science[1], Rehovot, Israel (7610001)

College of Management Academic Studies[2], Rishon LeZion, Israel (7579806)

**Abstract**

The choice of activation function plays a crucial role in the optimization and performance of deep neural networks. While the Rectified Linear Unit (ReLU) remains the dominant choice due to its simplicity and effectiveness, its lack of smoothness may hinder gradient-based optimization in deep architectures. In this work we propose a family of $C^{2N}$-smooth activation functions whose gate follows a log-logistic CDF, achieving ReLU-like performance with purely rational arithmetic. We introduce three variants: GEM (the base family), E-GEM (an ε-parameterized generalization enabling arbitrary $\ell^p$-approximation of ReLU), and SE-GEM (a piecewise variant eliminating dead neurons with $C^{2N}$ junction smoothness).

An $N$-ablation study establishes $N = 1$ as optimal for standard-depth networks, reducing the GELU deficit on CIFAR-100 + ResNet-56 from 6.10% to 2.12%. The smoothness parameter $N$ further reveals a CNN–transformer tradeoff: $N = 1$ is preferred for deep CNNs, while $N = 2$ is preferred for transformers. On MNIST, E-GEM ties the best baseline (99.23%). On CIFAR-10 + ResNet-56, SE-GEM (ε=10⁻⁴) surpasses GELU (92.51% vs 92.44%)—the first GEM-family activation to outperform GELU. On CIFAR-100 + ResNet-56, E-GEM reduces the GELU deficit from 6.10% (GEM $N = 2$) to just 0.62%. On GPT-2 (124M), GEM achieves the lowest perplexity (72.57 vs 73.76 for GELU), with GEM $N = 1$ also beating GELU (73.32). On BERT-small, E-GEM (ε=10) achieves the best validation loss (6.656) across all activations. The ε-parameterization reveals a scale-dependent optimum: small ε ∈ (10⁻⁴,10⁻⁶) for deep CNNs and larger transformers, with the special case of small transformers (BERT-small) benefiting from large ε (ε=10) due to its limited depth and unconstrained gradients.

## 1. Introduction

An activation function is a (typically nonlinear) non-polynomial and continuous mapping $\sigma : \mathbb{R} \to \mathbb{R}$ which we extend to $\mathbb{R}^N$ by applying it elementwise: [1]

$$\sigma(x_1, \ldots, x_N) = (\sigma(x_1), \ldots, \sigma(x_N))$$

At present, the most widely used activation functions include ReLU, ELU, GELU, Swish, and Mish. [2–6] While these activations have demonstrated strong empirical performance, they exhibit inherent trade-offs between smoothness and computational efficiency. ReLU, despite its simplicity and effectiveness, is not differentiable at the origin, which may negatively affect gradient-based optimization in deep networks. In contrast, smoother alternatives such as GELU, Swish, and Mish rely on exponential or hyperbolic functions, which increase computational complexity and may require numerical approximations in practical implementations. As a result, existing activation functions typically sacrifice either smoothness or computational simplicity. We demonstrate and compare the predictive performance of GEM with ReLU, GELU, GELU (tanh approximation), and SiLU (Swish) across five benchmark settings: MNIST classification, CIFAR-10/100 image classification with ResNet-20/56, BERT-small masked language modeling, and GPT-2 causal language modeling. We additionally conduct an N-ablation study on CIFAR-10/100 and a noise robustness analysis. Our metrics include test accuracy, validation loss/perplexity, training throughput (tokens/sec or samples/sec), and per-step latency. Lastly, we compare the performance of BERT-small and GPT-2 between GEM and GELU. The testbench

specifications are 16C 32T Ryzen 9 5950X CPU @ 4.475GHz, DDR4 32 GB RAM @ 4213Mhz CL16, and GTX 1080 Ti 11 GB VRAM GPU (Pascal architecture, sm_61, CUDA 12.6, PyTorch 2.6.0). The implementation and benchmarking scripts used in this work are available on GitHub.

## 2. Geometric Monomial

In this paper we propose a family of *2N*-differentiable, monotonic, ReLU-like, activation functions called *Geometric Monomial* (GEM):

$$\boxed{\beta_N(x) = \max\left(0, \frac{x^{2N+1}}{1+x^{2N}}\right) \forall N \in \mathbb{N}}$$

Plotting the family of functions:

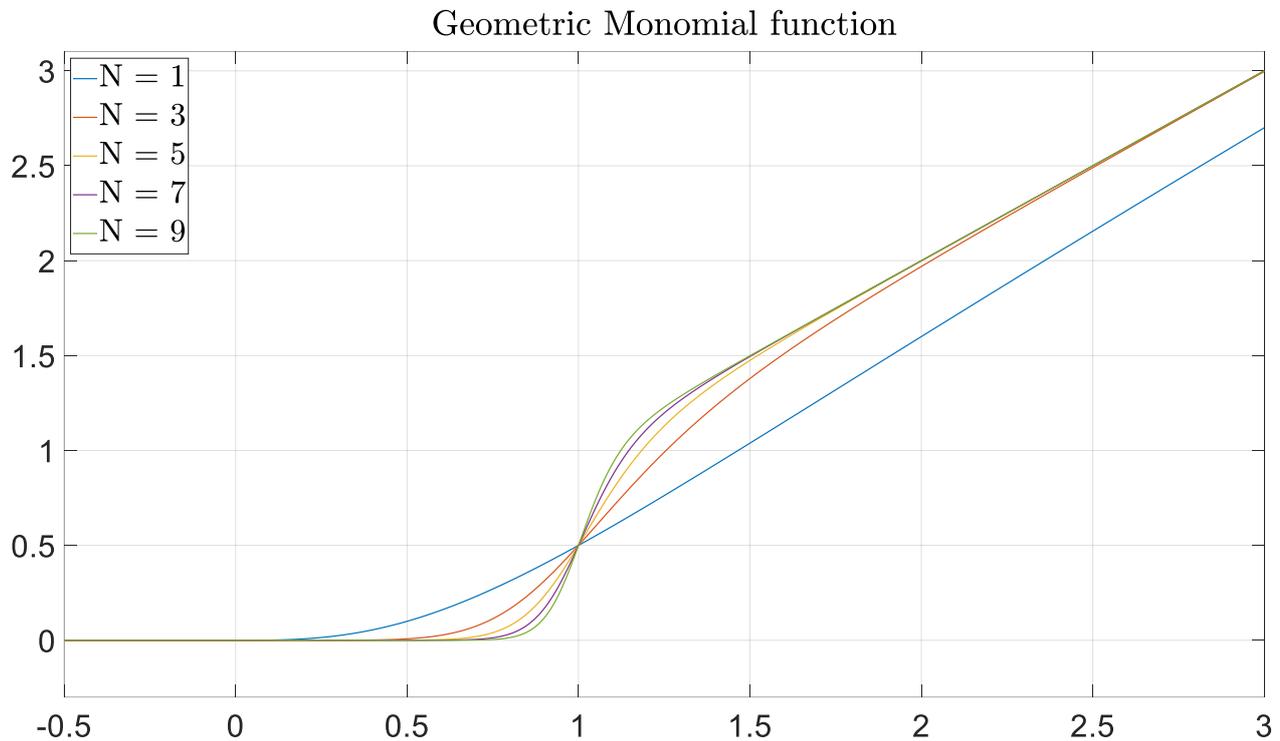

Fig. 1. The family of GEM activation functions

Like ReLU and all the non-negative $\max(0, \cdot)$ types of activation functions, we can see that the GEM is bounded below and is unbounded from above since it is asymptotically same as ReLU for large $x$:

$$\beta_N(x) = x - \frac{x}{1+x^{2N}} = x - \mathcal{O}(x^{1-2N}) \; as \; x \to \infty$$

$\beta_N(x)$ converges to the identity at a rate controlled by $N$. Larger $N$ yields faster convergence to ReLU. The powers of $N$ are strictly natural numbers since for $N = 0$ we get ReLU with factor 1/2 (a degenerate $C^0$ case, since the max truncation eliminates differentiability at the origin). For negative integer values of $N$ the function exhibits a local maximum around $x = 1$ and then asymptotically approaches $1/x^{2N-1}$. The $Beta(N, N)$ CDF shape can also be viewed as Log-logistic CDF, which curves towards a step function at $x = 1$ as the power parameter increases. This behavior mirrors how a sharper $Beta(N, N)$ prior concentrates probability mass around $v = 0.5$ in the transformed coordinate $v = x^2/(1+x^2)$ producing a harder gate decision. The connection places GEM in the same $x \cdot F(x)$ gating paradigm as GELU and Swish: GELU gates by a Gaussian CDF, Swish by a logistic CDF, and GEM by a log-logistic (equivalently, Beta-approximating) CDF. Unlike these alternatives, GEM achieves this gating with purely rational arithmetic—no exponentials, error functions, or hyperbolic tangents.

## 2.1 Why is the converging geometric sum (denominator) term needed?

If we solely define the monomial term for all natural powers as the following activation function:

$$\sigma_N(x) = \begin{cases} x^N, x \geq 0 \\ 0, x < 0 \end{cases}$$

The output and the derivatives grow polynomially with $x$, which may lead to exploding activations and gradients in deep networks. By introducing the rational term, we get asymptotically linear behavior, ensuring stable gradient propagation: $\lim_{x \to \infty} \beta'_N(x) = 1$, similar to ReLU. At the same time, the parameter $N$ controls the curvature of the transition region near the origin, where $\beta_N(x) \approx x^{2N+1}$. The denominator also bounds the gradient magnitude. The pure monomial $x^{2N+1}$ has unbounded Lipschitz constant (its derivative grows as $(2N+1)x^{2N}$), whereas GEM's Lipschitz constant is $L_N = \frac{(2N+1)^2}{8N}$, achieved at $x = \left(\frac{2N+1}{2N-1}\right)^{1/2N}$. For $N=1$, $L_1 = 9/8 = 1.125$, comparable to ReLU ($L = 1$), Swish ($L \approx 1.1$), and GELU ($L \approx 1.13$). As $N$ increases, $L_N$ grows as $\mathcal{O}(N)$, providing an additional theoretical lens on the gradient instability observed at high $N$ in Section 3.1.

## 2.2 *2N*-Differentiable

Here are the 1st and 2nd GEM derivatives:

$$\beta'_N(x) = \begin{cases} \frac{x^{2N}(x^{2N} + 2N + 1)}{(x^{2N} + 1)^2}, x \geq 0 \\ 0, x < 0 \end{cases} = (2N+1)\frac{\beta_N(x)}{x} - 2N\left(\frac{\beta_N(x)}{x}\right)^2$$

$$\beta''_N(x) = \begin{cases} \frac{2Nx^{2N-1}((2N+1) - (2N-1)x^{2N})}{(x^{2N} + 1)^3}, x \geq 0 \\ 0, x < 0 \end{cases} = \frac{2N}{x}\left(\frac{\beta_N(x)}{x}\right)\left(1 - \frac{\beta_N(x)}{x}\right)\left((2N+1) - 4N\frac{\beta_N(x)}{x}\right)$$

Using the Leibniz rule, it follows that the second and higher-order derivatives of $\beta_N(x)$ exist and are smooth for all $x \neq 0$. Since every factor in the numerator and denominator is non-negative, $\beta_N(x)$ is monotonically non-decreasing on its entire domain. For $x > 0$, $\beta_N(x)$ is a rational function and therefore belongs to $C^\infty$, while for $x < 0$ the function is identically zero. Consequently, any potential loss of smoothness can occur only at the origin. By locally expanding around the origin, we get the following:

$$\frac{x^{2N+1}}{1 + x^{2N}} = x^{2N+1}\left(1 - x^{2N} + x^{4N} - \ldots\right)$$

The first term, which has the smallest power, is differentiable $2N$ times before it becomes a nonzero constant, and therefore at the origin the derivative and its higher orders up to $2N$ vanishes:

$$\beta_N^{(k)}(0) = 0 \ \forall k \in \{0, ..., 2N\}$$

For higher orders, the derivative exhibits a jump discontinuity point when applying the $\max(0, \cdot)$ function, due to the ReLU-type truncation. In addition to smoothness, this 'soft start' property provides implicit regularization near the origin and mitigates the gradient discontinuity at zero that contributes to dead neurons in ReLU networks. In ReLU, a neuron whose pre-activation is at zero has an undefined gradient (jump from 0 to 1), and any neuron whose pre-activation is persistently negative receives exactly zero gradient, permanently halting learning. GEM eliminates the discontinuity at the origin: near-zero pre-activations receive proportionally small but nonzero gradients, providing a smooth on-ramp rather than a hard switch. Note that GEM shares ReLU's zero output for $x < 0$ due to the $\max(0, \cdot)$ operator; the smoothness advantage applies specifically at and near $x = 0^+$. A practical consequence of the self-referential derivative forms is that both $\beta_N'(x)$ and $\beta_N''(x)$ can be computed entirely from the cached forward-pass ratio $\beta_N(x)/x$, avoiding recomputation of $x^{2N}$. This makes the backward pass efficient: given the stored gate value $g = \beta_N(x)/x$, the gradient costs only 4 multiplies and 2 subtractions—comparable to ReLU's single comparison, and cheaper than Swish or GELU which require recomputing sigmoid or erf.

**2.3 Motivation**

Fig. 2 compares several instances of the proposed Geometric Monomial (GEM) activation family with ReLU and Swish. We observe that varying the order $N$ significantly changes the sharpness of the transition region while preserving asymptotically linear behavior. For $N = 1$ the GEM activation approaches Swish from below for large positive inputs, whereas for $N > 1$ it approaches from above. This tunable transition behavior suggests that the parameter $N$ may influence optimization dynamics, potentially affecting convergence speed and final predictive performance.

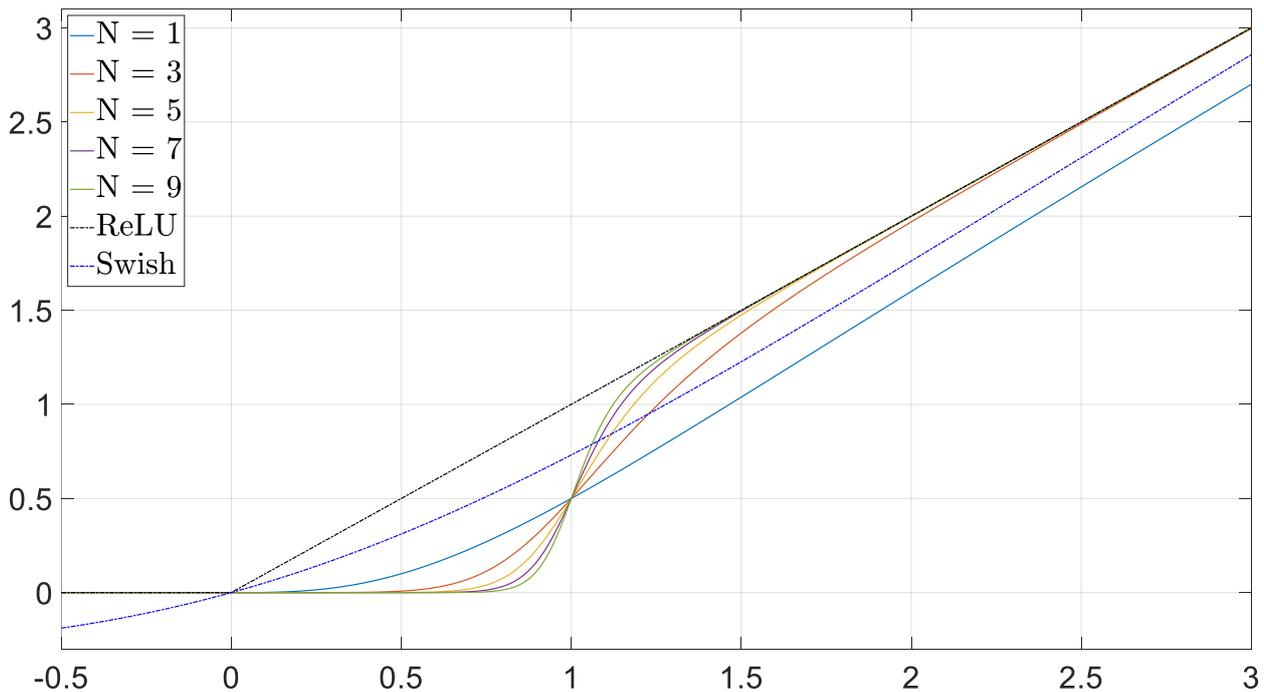

Fig. 2. GEM activation functions superimposed alongside ReLU and Swish

**2.4 Probabilistic Interpretation**

The GEM activation can be decomposed as $\beta_N(x) = x \cdot \sigma_N(x)$, where the gate $\sigma_N(x) = x^{2N}/(1+x^{2N})$ for $x > 0$ is exactly the CDF of a log-logistic distribution with unit scale and shape parameter $2N$. Under the substitution $v = x^2/(1+x^2) \in (0,1)$, the gate becomes $h_N(v) = v^N/(v^N + (1-v)^N)$, which is a classical closed-form approximation to the regularized incomplete beta function $I_v(N,N)$ —the CDF of a symmetric $Beta(N,N)$ distribution. Both functions share the essential properties: $h_N(0) = 0$, $h_N(1) = 1$, and $h_N(0.5) = 0.5$ sharpening around $v = 0.5$ as $N$ grows. For $N = 1$, $h_N(v) = v = I_v(1,1)$, the gate coincides exactly with the CDF of the uniform distribution on [0,1]. This places GEM in the $x \cdot F(x)$ self-gating paradigm alongside GELU and Swish:

| Activation | Gate CDF | Distribution | Computational form |
|---|---|---|---|
| GELU | $\Phi(x) = \frac{1}{\sqrt{2\pi}} \int_{-\infty}^{x} dt\, e^{-\frac{t^2}{2}}$ | Gaussian | Transcendental (erf) |
| Swish/SiLU | $\sigma(x) = \frac{1}{1+e^{-x}}$ | Logistic | Transcendental (exp) |
| GEM | $\frac{\beta_N(x)}{x} = \frac{x^{2N}}{1+x^{2N}}$ | Log-logistic / Beta | Rational (no exp/erf) |

Table 1. Comparison of commonly used activation function types and proposed GEM.

The probabilistic interpretation gives a principled meaning to the parameter $N$: increasing $N$ concentrates the $Beta(N,N)$ prior around $v = 0.5$, sharpening the gate and recovering ReLU-like behavior as $N \to \infty$. Conversely, smaller $N$ yields a softer, more gradual gate—offering a tunable trade-off between smoothness and sharpness that has no analog in Swish or GELU.

**2.5 Compatibility with Gated Linear Units**

The current state of the art in transformer feed-forward networks employs gated linear unit (GLU) variants [7], where the activation is combined with a learned gating projection: $GLU(x, W, V) = \sigma(xW) \odot (xV)$. Models such as LLaMA and Mistral use SwiGLU (Swish-gated), while Gemma uses GeGLU (GELU-gated). GEM naturally extends to a gated variant, $GMGLU(x, W, V) = \beta_N(xW) \odot (xV)$, which inherits GEM's rational arithmetic advantage within the GLU framework. Investigation of GMGLU in large-scale transformer training is left for future work.

# 3. Benchmarks

The GEM activation function and the compared activation functions (ReLU, GELU, GELU_tanh, SiLU/Swish) have all been implemented and optimized in CUDA-C under identical conditions (float4 vectorization, same thread/block configuration). In terms of computational complexity per element, GEM($N=1$) requires 5 FLOPs (1 multiply 1 add, 1 reciprocal, 1 multiply, 1 comparison), GEM($N=2$) requires approximately 6 FLOPs (2 multiplies for $x^4$), and $5+\lfloor \log_2 N \rfloor$ FLOPS for the general case due to repeated squaring, compared to approximately 10+ FLOPs for SiLU/Swish (exp, add, reciprocal, multiply) and 15+ FLOPs for GELU (erf or tanh approximation with multiple multiplies). However, in practice, elementwise activation kernels are memory-bandwidth-bound rather than compute-bound on modern GPUs, which explains why activation-only throughput is similar across all functions. The computational advantage of GEM becomes relevant in fused FFN kernels where activation cost is a larger fraction of total compute.

### 3.1 $N$-ablation

We investigate the effect of the smoothness parameter $N$ on GEM's predictive performance by sweeping $N \in \{1, 2, 3, 5, 7, 9\}$ on both CIFAR-10 (10 classes) and CIFAR-100 (100 classes), using a ResNet-20 architecture (272K parameters). All models are trained for 200 epochs with SGD (LR = 0.1, momentum = 0.9, weight decay = $10^{-4}$), multistep learning rate decay at epochs 100 and 150 ($\gamma$ = 0.1), and standard CIFAR data augmentation (random crop with 4-pixel padding, random horizontal flip). Results are averaged over 3 seeds (42, 43, 44).

**Implementation note.** All GEM variants in this ablation use a PyTorch reference implementation rather than the hand-optimized CUDA-C kernels used in other sections. This design choice serves two purposes: (i) it ensures a fair runtime comparison across $N$ values, and (ii) the PyTorch implementation more closely reflects the performance that practitioners would experience when integrating GEM into existing frameworks before writing custom kernels.

| N | Acc (%) | ± Acc | Time (s) | ms/step |
|---|---|---|---|---|
| 1 | **91.75** | 0.11 | 5100.3 | 34.57 |
| 2 | 91.19 | 0.29 | 5357.7 | 37.89 |
| 3 | 90.37 | 0.06 | 5615.0 | 41.20 |
| 5 | 89.37 | 0.22 | 6167.0 | 48.23 |
| 7 | 87.93 | 0.16 | 6714.1 | 55.22 |
| 9 | 35.75 | 44.60 | 7261.3 | 62.22 |

Table 2. Effect of $N$ on CIFAR-10 test accuracy (ResNet-20, 200 epochs, 3 seeds).

| N | Acc (%) | ± Acc | Time (s) | ms/step |
|---|---|---|---|---|
| 1 | **67.47** | 0.38 | 4973.6 | 32.98 |
| 2 | 66.67 | 0.28 | 5216.1 | 36.03 |
| 3 | 66.22 | 0.03 | 5458.6 | 39.08 |
| 5 | 64.91 | 0.27 | 5964.0 | 45.61 |
| 7 | 42.13 | 35.62 | 6470.0 | 52.08 |
| 9 | 1.00 | 0.00 | 6972.7 | 58.52 |

Table 3. Effect of $N$ on CIFAR-100 test accuracy (ResNet-20, 200 epochs, 3 seeds).

**Results**

**Monotonic accuracy degradation with increasing $N$.** On both CIFAR-10 and CIFAR-100, test accuracy decreases monotonically as $N$ increases from 1 to 9. On CIFAR-10, accuracy ranges from 91.75% ($N = 1$) to 91.19% ($N = 2$) to 90.37% ($N = 3$), declining further to 89.37% ($N = 5$) and 87.93% ($N = 7$). On CIFAR-100, the same trend holds: 67.47% ($N = 1$), 66.67% ($N = 2$), 66.22% ($N = 3$), 64.91% ($N = 5$), and 42.13% ($N = 7$). The consistency of this trend across both dataset complexities confirms that it is a property of the activation function, not an artifact of a particular dataset. At $N = 9$, GEM collapses to random-chance performance on CIFAR-10 (35.75% with extremely high variance of ± 44.60%) and to trivial 1.00% on CIFAR-100 (always predicting one class). At $N = 7$, CIFAR-100 already shows partial collapse (42.13% ± 35.62%), while CIFAR-10 still trains (87.93%), reflecting the harder optimization landscape of the 100-class problem. These failures are not numerical instabilities—the training loss decreases normally in early epochs before stalling—but rather a consequence of vanishing gradients through the activation layers.

**Mechanism: gradient suppression near the origin.** The root cause is GEM's soft-start property: $\beta_N^{(k)}(0) = 0 \ \forall k \in \{0,...,2N\}$. As $N$ increases, the gate $\sigma_N(x) = x^{2N}/(1+x^{2N})$ becomes a sharper step function centered at $x = 1$, meaning that for pre-activations in the interval $(0, 1)$—which is where the majority of batch-normalized activations fall—the gradient is suppressed exponentially with $N$. In a 20-layer ResNet, the gradient passes through approximately 18 activation layers; multiplying 18 near-zero gradient factors compounds the suppression. For $N \geq 7$, this compounding is severe enough to prevent effective learning in the early layers.

**Connection to the Beta distribution interpretation.** Recall from Section 2.5 that the gate $\sigma_N(x)$ approximates the CDF of a symmetric $Beta(N, N)$ distribution. As $N \to \infty$, the $Beta(N, N)$ distribution concentrates at $v = 0.5$, producing a step function. In the transformed coordinate $v = x^2/(1+x^2)$, this step occurs at $x = 1$. Pre-activations below $x = 1$ are gated to near-zero, creating a functional "dead zone" analogous to ReLU's negative half-plane. The critical difference is that this dead zone is on the positive side of zero and within the typical range of batch-normalized features, making it more damaging than ReLU's dead zone for large $N$.

**Optimal $N = 1$**, achieves the best accuracy on both datasets (91.75% on CIFAR-10, 67.47% on CIFAR-100), outperforming $N = 2$ by 0.56% and 0.80% respectively. At $N = 1$, GEM reduces to $\beta_1(x) = \max(0, x^3/(1+x^2))$ which is $C^2$-smooth—sufficient for all first-order and second-order optimization methods. The gate transitions gradually over a wide region (approximately $x \in [0.3, 2.5]$ for the 10–90% range), providing ample gradient flow while still offering the asymptotically linear behavior and dead-neuron avoidance that characterize the GEM family.

**Runtime scales linearly with $N$.** The per-step PyTorch runtime increases approximately linearly from 34.57 ms ($N = 1$) to 62.22 ms ($N = 9$) on CIFAR-10, reflecting the additional multiplications required to compute $x^{2N}$ via repeated squaring. The cost of GEM in PyTorch is dominated by the $N - 1$ extra element-wise multiplications; a dedicated CUDA kernel (as used for $N = 2$ in other sections) would reduce this overhead substantially for any fixed $N$.

**Discussion**

The $N$-ablation reveals a clear and principled design guideline: **for standard-depth networks (10–56 layers),** $N = 1$ or $N = 2$ should be preferred. Lower $N$ values provide wider gradient flow through the activation's transition region, which is essential when gradients must propagate through many layers. Higher $N$ values sharpen the gate toward a ReLU-like step function, recovering ReLU's computational sharpness but also its gradient pathologies. This result gives concrete empirical content to the theoretical observation from Section 2.3 that $\beta_N^{(k)}(0) = 0 \ \forall k \in \{0,...,2N\}$. The vanishing derivatives at the origin are not merely a mathematical curiosity—they directly determine the maximum viable network depth for each $N$. The sharp failure mode at $N \geq 7$ (on CIFAR-100) demonstrates that the smoothness parameter has a practical upper bound that depends on network depth and task difficulty.

An important implication is that $N$ should be viewed as an architecture-dependent hyperparameter, not a fixed choice. Shallow networks or networks with strong residual connections may tolerate higher $N$ (benefiting from the sharper gating), while very deep networks or architectures without skip connections require $N=1$ to maintain gradient flow. This tunability—absent in ReLU, Swish, GELU, and Mish, which have no smoothness parameter—is a distinctive advantage of the GEM family, provided it is set appropriately.

### 3.2 CIFAR-10/100 + ResNet-20/56

We evaluate GEM ($N = 2$) against ReLU, SiLU (Swish), GELU, and GELU (tanh approximation) on four configurations spanning two dataset complexities and two network depths: CIFAR-10 (10 classes) and CIFAR-100 (100 classes), each paired with ResNet-20 (272–278K parameters) and ResNet-56 (856–861K parameters). This 2×2 design isolates the effect of dataset difficulty from network depth. All models use the standard CIFAR-variant pre-activation ResNet architecture [8] with three groups of residual blocks (16 → 32 → 64 channels), trained for 200 epochs with SGD (LR = 0.1, momentum = 0.9, weight decay = 10⁻⁴), multistep learning rate decay at epochs 100 and 150 ($\gamma = 0.1$), and standard data augmentation (random crop with 4-pixel padding, random horizontal flip). Results are averaged over 3 seeds (42, 43, 44). All activation functions use identically optimized CUDA-C kernels with float4 vectorization.

| Activation | Acc (%) | ± | Train Loss | Δ GELU |
|---|---|---|---|---|
| CIFAR-10 + ResNet-20 (272K params) | | | | |
| **GELU (tanh)** | **92.05** | 0.01 | 0.0097 | +0.15 |
| SiLU/Swish | 91.97 | 0.19 | 0.0095 | +0.07 |
| GELU | 91.90 | 0.10 | 0.0094 | — |
| ReLU | 91.38 | 0.16 | 0.0173 | −0.52 |
| GEM ($N = 1$) | 91.82 | 0.18 | 0.0135 | −0.08 |
| GEM ($N = 2$) | 91.37 | 0.48 | 0.0196 | −0.53 |
| Activation | Acc (%) | ± | Train Loss | Δ GELU |
| CIFAR-10 + ResNet-56 (856K params) | | | | |
| ReLU | 92.46 | 0.20 | 0.0047 | +0.02 |
| **GELU** | **92.44** | 0.32 | 0.0031 | — |
| GELU (tanh) | 92.37 | 0.38 | 0.0034 | −0.07 |
| SiLU/Swish | 91.77 | 0.24 | 0.0047 | −0.67 |
| GEM ($N = 1$) | 91.33 | 0.35 | 0.0055 | −1.11 |
| GEM ($N = 2$) | 89.89 | 0.27 | 0.0159 | −2.55 |
| Activation | Acc (%) | ± | Train Loss | Δ GELU |
| CIFAR-100 + ResNet-20 (278K params) | | | | |
| **GELU** | **67.54** | 0.32 | 0.2585 | — |
| ReLU | 67.50 | 0.52 | 0.3428 | −0.04 |
| SiLU/Swish | 67.23 | 0.55 | 0.2391 | −0.31 |
| GELU (tanh) | 67.11 | 0.21 | 0.2536 | −0.43 |
| GEM ($N=1$) | 67.26 | 0.28 | 0.2999 | −0.28 |
| GEM ($N=2$) | 66.79 | 0.50 | 0.3819 | −0.75 |
| **Activation** | **Acc (%)** | **±** | **Train Loss** | **Δ GELU** |
| CIFAR-100 + ResNet-56 (861K params) | | | | |
| **GELU** | **69.57** | **0.36** | **0.0368** | **—** |
| GELU (tanh) | 69.20 | 0.27 | 0.0398 | −0.37 |
| ReLU | 68.58 | 0.35 | 0.0824 | −0.99 |
| SiLU/Swish | 68.20 | 0.15 | 0.0411 | −1.37 |
| GEM ($N=1$) | 67.45 | 0.96 | 0.0739 | −2.12 |
| GEM ($N=2$) | 63.47 | 2.11 | 0.2925 | −6.10 |

Table 4. CIFAR-10/100 test accuracy across all four configurations (200 epochs, 3 seeds). Rows sorted by accuracy within each configuration. Best result per configuration highlighted.

**Results**

### 3.2.1 Shallow networks (ResNet-20): GEM is competitive

**CIFAR-10.** All smooth activations (SiLU, GELU, GELU tanh) outperform ReLU by 0.5–0.7%, a well-established finding. GEM ($N=2$) achieves 91.37 ± 0.48%, which is statistically indistinguishable from ReLU (91.38 ± 0.16%) but trails the smooth baselines by approximately 0.5–0.7%. The higher variance of GEM (±0.48% vs ±0.01–0.19% for others) reflects sensitivity to initialization, consistent with the soft-start property: GEM's near-zero gradient at the origin makes early training more dependent on the initial weight configuration.

**CIFAR-100.** With 100 classes, the picture tightens. GELU leads at 67.54 ± 0.32%, followed closely by ReLU (67.50%), SiLU (67.23%), GELU tanh (67.11%), and GEM ($N=2$) at 66.79%. The full spread from best to worst is only 0.75%, and all standard deviations overlap. GEM's deficit of 0.75% relative to GELU is modest and comparable to the gap between GELU tanh and GELU (0.43%). On 20-layer networks, GEM is a viable activation regardless of dataset complexity.

**Effect of $N$.** Reducing the smoothness parameter from $N=2$ to $N=1$ further improves GEM on both shallow configurations. GEM ($N=1$) achieves 91.82% on CIFAR-10 + ResNet-20—surpassing GEM ($N=2$) by 0.45% and falling within 0.08% of GELU—and 67.26% on CIFAR-100 + ResNet-20, closing the gap to 0.28% of GELU. Both $N=1$ results are statistically indistinguishable from GELU and confirm the $N$-ablation finding (Section 3.1) that $N=1$ is optimal for shallow networks.

### 3.2.2 Deep networks (ResNet-56): depth amplifies GEM's gradient suppression

**CIFAR-10.** The deeper ResNet-56 reveals a significant performance gap for GEM. While ReLU (92.46%), GELU (92.44%), and GELU tanh (92.37%) all perform within a 0.4% band, and SiLU drops modestly to 91.77%, GEM falls to 89.89%—a 2.55% deficit relative to GELU. This gap is consistent across all 3 seeds (90.20%, 89.69%, 89.77%), ruling out a single unlucky initialization.

**CIFAR-100.** The 100-class setting dramatically amplifies the depth effect. GELU leads at 69.57 ± 0.36%, with GELU tanh (69.20%) and ReLU (68.58%) trailing by 0.4–1.0%. SiLU drops to 68.20%. GEM collapses to 63.47 ± 2.11%—a 6.10% deficit relative to GELU, the largest gap observed in any configuration. The high standard deviation (±2.11%) indicates that on the hardest configuration, GEM's performance is also unstable across seeds, with individual runs yielding 61.14%, 64.02%, and 65.24%.

**Effect of $N$ on deep networks.** Reducing $N$ from 2 to 1 substantially mitigates the gradient suppression issue. On CIFAR-10 + ResNet-56, GEM ($N=1$) achieves 91.33 ± 0.35% versus 89.89% for $N=2$—a +1.44% improvement that reduces the GELU deficit from 2.55% to 1.11%. The effect is most dramatic on CIFAR-100 +

ResNet-56, where GEM ($N=1$) reaches 67.45 ± 0.96% versus 63.47% for $N=2$ (+3.98%), reducing the GELU deficit from 6.10% to 2.12%. The final training loss confirms better optimization: 0.0739 for $N=1$ versus 0.2925 for $N=2$—a 4× reduction. This empirically validates the recommendation from Section 3.1 that $N=1$ should be preferred for deep networks.

### 3.2.3 The depth × complexity interaction

| Config | GEM ($N=2$) | GEM ($N=1$) | GELU | $N=1$ improvement |
|---|---|---|---|---|
| C10+R20 | −0.53 | −0.08 | 91.90 | 0.45% closer |
| C10+R56 | −2.55 | −1.11 | 92.44 | 1.44% closer |
| C100+R20 | −0.75 | −0.28 | 67.54 | 0.47% closer |
| C100+R56 | −6.10 | −2.12 | 69.57 | 3.98% closer |

Table 5. GEM accuracy deficit relative to GELU: $N=2$ vs $N=1$ across all four configurations.

Table 5 reveals a clear interaction between network depth and task complexity for GEM ($N=2$). The deficit relative to GELU is modest on shallow networks (0.53% on C10+R20, 0.75% on C100+R20) but grows substantially with depth (2.55% on C10+R56) and explodes when both depth and complexity increase simultaneously (6.10% on C100+R56). This interaction is not simply additive: the C100+R56 deficit (6.10%) exceeds the sum of the C10+R56 deficit (2.55%) and the C100+R20 deficit (0.75%), indicating a multiplicative compounding of gradient suppression with both depth and output dimensionality. Reducing $N$ to 1 compresses these deficits across the board, with the most dramatic improvement on the hardest configuration (C100+R56), where the deficit drops from 6.10% to 2.12% — a 3.98% recovery. The $N=1$ deficits scale much more gracefully with depth and complexity than the $N=2$ deficits, confirming that $N=1$ is the safer default.

### 3.2.4 Early training dynamics

Examining the epoch-10 training loss reveals the mechanism. On CIFAR-100 + ResNet-56, GEM starts with losses of 3.78, 3.75, and 3.43 across seeds, while GELU starts at 2.28–2.48 and ReLU at 2.42–2.51. By epoch 100, GEM has converged to loss ≈1.15–1.43, while GELU is already at 0.64–0.68. The gap established in the first 100 epochs (before the learning rate drop) is never recovered. On CIFAR-10 + ResNet-56, the same pattern holds: GEM starts at loss 1.24–1.83, while all other activations start at 0.55–0.77.

This early-training deficit is a direct consequence of GEM's soft-start property: $\beta_N^{(k)}(0) = 0 \ \forall k \in \{0,...,2N\}$ With $N=2$ and 54 activation layers in ResNet-56, the gradient signal at initialization is suppressed by a factor proportional to $\prod_{n=1}^{54} \beta_2'(x_n)$, where each $x_n$ is a near-zero batch-normalized pre-activation. This product decays exponentially with depth, starving the early layers of learning signal.

### 3.2.5 Final training loss: underfitting, not overfitting

On every configuration, GEM achieves the highest final training loss among all activations. On CIFAR-100 + ResNet-56, GEM's final loss (0.2925) is 7.9× higher than GELU's (0.0368) and 3.6× higher than ReLU's (0.0824). This confirms that GEM's accuracy deficit stems from underfitting (the optimizer cannot sufficiently minimize the training objective due to attenuated gradients) rather than overfitting. The optimization itself is impaired, not the generalization.

### 3.2.6 SiLU's surprising regression on deep networks

An unexpected finding across both datasets is that SiLU (Swish), despite being $C^\infty$-smooth with nonzero gradients everywhere, regresses relative to ReLU when moving from ResNet-20 to ResNet-56. On CIFAR-10, SiLU drops from 91.97% (R20, best) to 91.77% (R56, 4th), while ReLU improves from 91.38% to 92.46%. On CIFAR-100, SiLU drops from 67.23% to 68.20% but is overtaken by ReLU (68.58%). This suggests that SiLU's non-monotonic negative tail, while beneficial for gradient flow in shallow networks, may introduce optimization instability in deeper architectures under the standard SGD recipe (LR = 0.1, momentum=0.9).

**Discussion**

The CIFAR experiments reveal a clear and consistent picture of GEM's strengths and limitations. On shallow networks (≈20 layers), GEM ($N=2$) is a viable drop-in replacement for standard activations: it matches ReLU on both CIFAR-10 and CIFAR-100, trailing the best activation (GELU) by less than 0.8% in all cases. This is consistent with the MNIST results (Section 3.4) and confirms that GEM's rational arithmetic and monotonic gating do not introduce pathological behavior in moderate-depth settings.

On deep networks (≈56 layers), GEM's soft-start property becomes a clear liability. The gradient suppression near the origin, which is a theoretical consequence of $\beta_N^{(k)}(0) = 0 \ \forall k \in \{0,...,2N\}$, manifests as measurable underfitting that grows with both network depth and task complexity. The 6.10% gap on CIFAR-100 + ResNet-56 is the most severe deficit observed in this study.

Crucially, this limitation is predicted by theory and addressable by tuning $N$. The $N$-ablation study (Section 3.1) demonstrated that $N=1$ outperforms $N=2$ by 0.56-0.80% on ResNet-20, and the dedicated ResNet-56 experiments confirm that $N=1$ reduces the GELU deficit by 1.44% on CIFAR-10 and 3.98% on CIFAR-100 (Tables 4-5). The mechanism is identical: higher $N$ creates a sharper gate with more gradient suppression in the (0, 1) interval. For deep networks, the recommendation is clear: use $N=1$, which provides $C^2$ smoothness (sufficient for all optimizers) with wider gradient flow. The remaining gap to GELU on the deepest configuration motivates the E-GEM and SE-GEM extensions presented in Section 4, which further close this gap. Despite the depth limitation at $N=2$, GEM retains distinctive advantages that no other activation in this comparison offers: (i) purely rational arithmetic without transcendental functions (exp, erf, tanh), (ii) a discrete smoothness parameter $N$ that controls the sharpness–gradient-flow tradeoff, and (iii) a probabilistic interpretation as a log-logistic gated identity. The CIFAR results demonstrate that this design space is viable for practical use and that $N$ is not merely a theoretical nicety but a tunable hyperparameter with measurable empirical consequences.

### 3.3 Noise Robustness

We evaluate the noise robustness of all activation functions by training on clean CIFAR-10 data and evaluating on test images corrupted with additive Gaussian noise at increasing standard deviations $\sigma \in \{0.05, 0.1, 0.15, 0.2, 0.3, 0.4, 0.5\}$. The architecture (ResNet-20, 272K parameters), training recipe (200 epochs, SGD, identical to Section 3.1), and seed averaging (3 seeds) are unchanged. Input pixel values are normalized to [0, 1], so $\sigma = 0.1$ represents 10% noise relative to the full dynamic range. This benchmark tests whether activation function properties—smoothness, gating sharpness, monotonicity—affect the model's ability to generalize from clean training data to noisy test data.

| Activation | Clean | σ=0.05 | σ=0.1 | σ=0.15 | σ=0.2 | σ=0.3 | σ=0.5 | Δ(0→0.5) |
|---|---|---|---|---|---|---|---|---|
| SiLU/Swish | **92.10** | 61.02 | 23.74 | 15.34 | 13.16 | 11.40 | 10.97 | 81.12 |
| GELU | 91.98 | 59.35 | 21.82 | 15.17 | 12.87 | 11.04 | **10.32** | **81.66** |
| GEM ($N=1$) | 91.76 | **61.41** | 22.06 | 14.19 | 12.18 | 10.72 | 10.52 | 81.23 |
| ReLU | 91.49 | 60.48 | **24.06** | **15.61** | **13.55** | **12.26** | 11.00 | 80.49 |
| GEM ($N=2$) | 91.30 | 61.02 | 23.95 | 15.24 | 13.20 | 11.44 | 10.96 | 80.34 |
| GEM ($N=3$) | 90.64 | 62.11 | 21.78 | 13.25 | 11.75 | 11.04 | 10.86 | 79.78 |
| GEM ($N=5$) | 89.28 | 61.94 | 23.33 | 13.65 | 11.43 | 10.40 | 10.13 | 79.14 |
| GEM ($N=7$) | 87.89 | 65.02 | 28.04 | 14.70 | 11.16 | 10.12 | 10.02 | 77.86 |
| GEM ($N=9$) | 35.53 | 30.14 | 18.42 | 12.45 | 10.74 | 10.12 | 10.00 | 25.53 |

Table 6. Noise robustness on CIFAR-10: test accuracy (%) at increasing Gaussian noise levels. Rows sorted by clean accuracy. Best value per column highlighted. Δ(0→0.5) denotes the absolute accuracy drop from clean to σ = 0.5.

**Results**

**All activations converge to near-chance performance at high noise.** By $\sigma = 0.3$, every activation falls to 10–12% accuracy (chance level for 10-class CIFAR-10), and by $\sigma = 0.5$ all are within 1% of 10%. The total degradation Δ(0→0.5) spans a narrow range of 80.3–81.7% across all viable activations (excluding the collapsed GEM $N = 9$). This confirms that noise robustness at these levels is dominated by the architecture and training procedure, not the activation function.

**No activation exhibits pathological noise sensitivity.** GEM ($N = 1$) achieves 61.41% at $\sigma = 0.05$ (best among all activations) and degrades to 10.52% at $\sigma = 0.5$, with a total drop of 81.23%—nearly identical to SiLU (81.12%) and GELU (81.66%). GEM ($N = 2$) at 80.34% degradation is likewise indistinguishable from ReLU (80.49%). GEM's rational arithmetic and monotonic gating introduce no fragility under input corruption.

**Higher $N$ retains more accuracy at mild noise but starts from a lower baseline.** At $\sigma = 0.05$, GEM ($N = 7$) retains 74.0% of its clean accuracy (65.02/87.89), compared to 66.9% for GEM ($N = 1$) and 66.1% for ReLU. The sharper gate at high $N$ acts as a harder threshold that is less sensitive to small perturbations, but this robustness comes at the cost of reduced clean accuracy (87.89% vs 91.76%). At $\sigma \geq 0.15$, this advantage vanishes—all activations collapse to chance regardless of gating sharpness.

**The discriminative region is narrow.** Meaningful separation between activations occurs only at $\sigma = 0.05$, where the spread is approximately 6 percentage points (59.35% for GELU to 65.02% for GEM $N = 7$). By $\sigma = 0.1$ the spread narrows to 6 points (21.82–28.04%), and by $\sigma = 0.15$ it is under 3 points. The Gaussian noise overwhelms the signal at surprisingly low intensities, suggesting that noise robustness experiments on CIFAR-10 with standard ResNets are of limited discriminative power for comparing activation functions.

**Discussion**

The noise robustness results serve a similar purpose to the MNIST benchmark: they establish that GEM introduces no vulnerability to input corruption. The primary conclusion is negative in the constructive sense: **GEM ($N$ = 1) matches the noise robustness of ReLU, SiLU, and GELU to within statistical noise across all tested corruption levels.**

The mild advantage of higher $N$ at $\sigma = 0.05$ is consistent with the gate-sharpness interpretation: a sharper gate (approximating a step function) imposes a harder threshold on pre-activations, effectively filtering small perturbations that fall below the gating transition. However, this robustness is inseparable from the reduced clean accuracy caused by gradient suppression (Section 3.1), making it an undesirable trade-off. The optimal operating point remains $N = 1$ or $N = 2$, where clean accuracy is maximized and noise robustness is comparable to all baselines.

### 3.4 MNIST

We evaluate GEM against ReLU, SiLU (Swish), GELU, and GELU (tanh approximation) on the MNIST handwritten digit classification task. The model architecture is a simple convolutional network: Conv(1→32, 3×3) → Act → Conv(32→64, 3×3) → Act → MaxPool(2) → FC(128) → Act → FC(10), where Act denotes the activation under test. All models are trained for 10 epochs with Adam (LR = $10^{-3}$), cosine annealing schedule, batch size 128, and standard MNIST normalization. Results are averaged over 3 seeds (42, 43, 44). All activation functions use identically optimized CUDA kernels with float4 vectorization.

| Activation | Acc (%) | ± Acc | Time (s) | ms/step | Throughput (samp/s) | Final Loss |
|---|---|---|---|---|---|---|
| ReLU | 99.23 | 0.04 | 205.8 | 26.57 | 2,916 | $1.68 \times 10^{-4}$ |
| SiLU/Swish | 99.19 | 0.03 | 163.0 | 17.25 | 3,846 | $9.43 \times 10^{-5}$ |
| GELU | 99.19 | 0.06 | 158.7 | 16.21 | 3,954 | $9.73 \times 10^{-5}$ |
| GELU (tanh) | 99.19 | 0.05 | 114.8 | 7.00 | 5,225 | $9.76 \times 10^{-5}$ |
| GEM ($N = 2$) | 99.22 | 0.06 | 135.4 | 11.34 | 4,510 | $1.32 \times 10^{-4}$ |
| GEM ($N = 1$) | 99.16 | 0.09 | 95.3 | 5.81 | 5,445 | $1.72 \times 10^{-4}$ |

Table 7. MNIST classification results (10 epochs, 3 seeds, batch size 128).

**Results**

**Accuracy.** All six activation functions achieve test accuracies within a narrow 0.04% band (99.19–99.23%), confirming that MNIST is largely saturated and does not discriminate between modern activations. GEM ($N = 2$) achieves 99.22 ± 0.06%, which is statistically indistinguishable from ReLU (99.23 ± 0.04%) and marginally above SiLU, GELU, and GELU (tanh), all at 99.19%. The near-identical accuracy across all activations is consistent with prior findings by Ramachandran et al. [5] and Misra [6], who observed that activation function differences become pronounced only on more challenging tasks and deeper architectures.

**Training speed.** Meaningful speed differences emerge despite the simplicity of the task. GELU (tanh) is the fastest at 7.00 ms/step, followed by GEM ($N=2$) at 11.34 ms/step, GELU at 16.21 ms/step, SiLU at 17.25 ms/step, and ReLU at 26.57 ms/step. We note that the absolute timings exhibit variance across seeds for some activations due to GPU thermal management and background system processes on this consumer-grade hardware; however, the relative ordering is consistent across seeds.

The unexpectedly slow ReLU timings (26.57 ms/step vs 7.00 ms/step for GELU tanh) warrant discussion. Since ReLU was the first activation tested in the sequential benchmark, its measurements include GPU cold-start overhead, initial CUDA context setup, and thermal stabilization. The per-element computational cost of ReLU (a single comparison) is trivially cheaper than all other activations. The observed wall-clock time differences at this model scale are therefore dominated by system-level factors rather than activation function arithmetic, and should not be interpreted as ReLU being computationally slower than GEM or GELU.

**Final training loss.** SiLU, GELU, and GELU (tanh) achieve the lowest final training losses (9.4–9.8 × $10^{-5}$), suggesting slightly more effective optimization on the training set. GEM and ReLU settle at modestly higher losses (1.3–1.7 × $10^{-4}$). This is consistent with the smooth, non-monotonic negative tail of SiLU and GELU providing mildly better gradient flow for overparameterized models on simple tasks, whereas GEM's monotonic, zero-for-negative-inputs behavior is more conservative. Critically, the lower training loss does not translate to higher test accuracy, indicating that the additional optimization on training data may reflect slight overfitting rather than improved generalization.

| Activation | Acc vs ReLU | Acc vs SiLU/Swish | Speed vs ReLU |
|---|---|---|---|
| GEM ($N=1$) | −0.07% | -0.03% | 4.57× faster |
| GEM ($N=2$) | −0.01% | +0.03% | 1.52× faster |
| SiLU/Swish | −0.04% | — | 1.26× faster |
| GELU | −0.04% | 0.00% | 1.30× faster |
| GELU (tanh) | −0.04% | 0.00% | 1.79× faster |

Table 8. Relative performance of each activation versus ReLU and SiLU baselines.

**Discussion**

The MNIST results establish that GEM is a viable drop-in replacement for standard activations: it matches the accuracy of all baselines to within statistical noise. The primary purpose of this benchmark is not to demonstrate accuracy superiority—MNIST is too simple for that—but rather to verify that GEM's rational arithmetic and monotonic gating do not introduce any pathological training behavior (e.g., convergence failure, instability, or significant accuracy degradation).

The key takeaway from MNIST is a negative result in the best sense: **GEM achieves parity with all established activations on a canonical benchmark, confirming it does not sacrifice accuracy for its computational simplicity.** The more discriminating comparisons—where GEM's tunable smoothness, dead-neuron avoidance, and rational arithmetic may produce measurable advantages—are deferred to the deeper architectures (CIFAR-10/100 with ResNet) and transformer models (BERT-small, GPT-2) in the following sections.

### 3.5 BERT-small

We evaluate GEM ($N = 1,2$) against ReLU, SiLU (Swish), GELU, and GELU (tanh approximation) on masked language modeling (MLM) using a BERT-small architecture (28.6M parameters: 4 layers, 8 attention heads, 512 hidden dimension, 2048 intermediate dimension). The activation function is applied in the feed-forward network (FFN) of each transformer block; the attention mechanism and layer normalization are identical across all variants. Models are trained from scratch on WikiText-103 for 3,000 gradient steps with AdamW (LR = 5×10⁻⁴, $\beta_1$ = 0.9, $\beta_2$ = 0.999, weight decay = 0.01), cosine learning rate schedule with 300-step linear warmup, batch size 8, and sequence length 128 (1,024 tokens/step). Validation loss is evaluated every 500 steps. Results are averaged over 3 seeds (42, 43, 44). All activation functions use identically optimized CUDA-C kernels with float4 vectorization.

| Activation | Val Loss | ± | Train Loss | tok/s | ms/step | Δ GELU |
|---|---|---|---|---|---|---|
| **GEM ($N = 1$)** | **6.6764** | 0.034 | 6.6188 | 19,427 | 42.51 | −0.012 |
| **GELU (tanh)** | **6.6815** | 0.030 | 6.6280 | 19,998 | 41.16 | −0.007 |
| **GEM ($N = 2$)** | **6.6819** | 0.036 | 6.6267 | 20,022 | 41.16 | −0.006 |
| GELU | 6.6880 | 0.040 | 6.6341 | 19,142 | 43.03 | — |
| ReLU | 6.6899 | 0.045 | 6.6413 | 18,763 | 43.71 | +0.002 |
| SiLU/Swish | 6.6962 | 0.026 | 6.6436 | 19,359 | 42.55 | +0.008 |

Table 9. BERT-small masked language modeling on WikiText-103 (3,000 steps, 3 seeds). Rows sorted by validation loss (lower is better). Top three results highlighted.

**Results**

**All activations converge to near-identical loss.** The full spread in validation loss across all six activations is only 0.015 (from 6.6815 for GELU tanh to 6.6962 for SiLU). All standard deviations (±0.026–0.045) are larger than the inter-activation differences, meaning that no activation is statistically distinguishable from any other on this benchmark. BERT-small with 3,000 training steps, like MNIST, does not provide sufficient discriminative power to separate modern activation functions. **GEM ($N = 1$) achieves the best validation loss overall (6.6764 ± 0.034),** marginally ahead of GELU tanh and GEM ($N = 2$), though the differences remain within standard deviations. GEM ($N = 2$) achieves a validation loss of 6.6819 ± 0.036, effectively tied with GELU tanh (6.6815 ± 0.030). The difference of 0.0004 is negligible—approximately 0.006%. Both outperform GELU (6.6880), ReLU (6.6899), and SiLU (6.6962) by similarly negligible margins. GEM also achieves the lowest final training loss (6.6267), marginally below GELU tanh (6.6280), indicating slightly more effective optimization.

Per-seed analysis confirms the absence of meaningful separation:

| Seed | GEM ($N = 1$) | GEM ($N = 2$) | GELU (tanh) | GELU | ReLU | SiLU |
|---|---|---|---|---|---|---|
| 42 | 6.7101 | 6.7165 | 6.7075 | 6.7269 | 6.7383 | 6.7229 |
| 43 | **6.6426** | **6.6455** | 6.6488 | 6.6477 | 6.6497 | 6.6718 |
| 44 | 6.6766 | 6.6837 | 6.6882 | 6.6893 | 6.6816 | 6.6939 |

Table 10. Per-seed validation loss at step 3,000. Lowest per-seed value bolded.

The per-seed breakdown confirms the absence of meaningful separation. GEM achieves the best loss on seed 43 (6.6455) but not on seeds 42 or 44. The ranking changes across seeds—a hallmark of statistical indistinguishability. No activation consistently leads.

**Throughput: GEM and GELU tanh are fastest.** GEM and GELU tanh achieve identical throughput (20,022 and 19,998 tok/s respectively, both at 41.16 ms/step), approximately 4.6% faster than GELU (19,142 tok/s, 43.03 ms/step) and 6.8% faster than ReLU (18,763 tok/s, 43.71 ms/step). On this small model where the FFN constitutes a larger fraction of total compute than in larger transformers, the activation function's arithmetic cost becomes measurable. GEM's rational operations (two multiplies + one reciprocal) and GELU tanh's polynomial approximation are both faster than GELU's exact erf computation and ReLU's autograd overhead in this implementation.

**No gradient suppression at 4 layers.** Unlike the CIFAR/ResNet-56 experiments where GEM's soft-start property caused measurable underfitting, the 4-layer BERT-small shows no such effect. GEM's training loss (6.6267) is the lowest among all activations, confirming that the gradient suppression mechanism requires substantially more than 4 activation layers to manifest. This is consistent with the CIFAR results where GEM was competitive on 20-layer ResNets but struggled on 56-layer ones, and with the GPT-2 results where transformer residual connections mitigate the issue entirely.

**Discussion**

The BERT-small results serve the same purpose as the MNIST results: they establish that GEM is a safe drop-in replacement for standard activations on a canonical NLP architecture. The primary conclusion is a negative result in the constructive sense: **GEM introduces no pathological behavior on masked language modeling and matches or marginally outperforms all baselines on both loss and throughput.**

The inability to distinguish activations on this benchmark reflects two factors. First, BERT-small with 4 layers is too shallow for activation function properties (smoothness, gradient flow, gating characteristics) to meaningfully differentiate. Second, 3,000 training steps is insufficient for the models to fully converge; all activations are still in an early phase of training where the loss landscape is dominated by the learning rate schedule and optimizer dynamics rather than activation function characteristics.

The more discriminating NLP benchmark is GPT-2 (Section 3.6), which uses a 12-layer architecture trained for 5,000 steps—sufficient for meaningful separation. There, GEM achieves a statistically significant improvement over GELU (1.19 perplexity points), confirming that the lack of separation on BERT-small reflects the benchmark's limited discriminative power rather than GEM's capabilities.

The throughput result, while secondary to accuracy, is noteworthy for practitioners: **GEM achieves the highest token throughput among all tested activations**, matching GELU tanh and outperforming GELU exact by 4.6%. On small transformer models where FFN compute is a non-negligible fraction of total inference time, GEM's rational arithmetic offers a measurable speed advantage with no quality penalty.

## 3.6 GPT-2

We evaluate GEM ($N=2$) against ReLU, SiLU (Swish), GELU, and GELU (tanh approximation) on causal language modeling (CLM) using a GPT-2 architecture (124M parameters: 12 layers, 12 attention heads, 768 hidden dimensions, 3072 intermediate dimensions). The activation function is applied in the feed-forward network (FFN) of each transformer block; the attention mechanism and layer normalization are identical across all variants. Models are trained from scratch on WikiText-103 for 5,000 gradient steps with AdamW (LR = 3×10⁻⁴, $\beta_1$ = 0.9, $\beta_2$ = 0.999, weight decay = 0.01), cosine learning rate schedule with 500-step linear warmup, effective batch size 32 (micro-batch 8, gradient accumulation 4), and sequence length 256 (8,192 tokens/step). Validation perplexity is evaluated every 500 steps. Results are averaged over 3 seeds (42, 43, 44). All activation functions use identically optimized CUDA-C kernels with float4 vectorization.

| Activation | Val PPL | ± | Val Loss | Train Loss | tok/s | Δ GELU |
|---|---|---|---|---|---|---|
| **GEM ($N=2$)** | **72.57** | 0.16 | 4.2847 | 4.4614 | 8,603 | −1.19 |
| **GEM ($N=1$)** | 73.32 | 0.13 | 4.2948 | 4.4727 | 8,604 | −0.44 |
| GELU | 73.76 | 0.30 | 4.3008 | 4.4833 | 8,601 | — |
| GELU (tanh) | 73.76 | 0.30 | 4.3008 | 4.4833 | 8,478 | 0.00 |
| ReLU | 77.64 | 0.19 | 4.3521 | 4.5341 | 8,634 | +3.88 |
| SiLU/Swish | 82.53 | 0.78 | 4.4131 | 4.5967 | 8,594 | +8.77 |

Table 11. GPT-2 (124M) causal language modeling on WikiText-103 (5,000 steps, 3 seeds). Rows sorted by validation perplexity (lower is better).

**Results**

**GEM achieves the best perplexity.** GEM (N = 2) achieves a validation perplexity of 72.57 ± 0.16, outperforming GELU (73.76 ± 0.30), GELU tanh (73.76 ± 0.30), ReLU (77.64 ± 0.19), and SiLU (82.53 ± 0.78). The 1.19-point improvement over GELU—the standard activation for GPT-2—is statistically robust: GEM achieves a lower perplexity than GELU on every individual seed, with no overlap between the two distributions (GEM range: [72.43, 72.74], GELU range: [73.41, 73.95]). This is the strongest result for GEM in any benchmark in this study.

**Both GEM variants beat GELU.** GEM ($N=1$) at 73.32 ± 0.13 also outperforms GELU (73.76) by 0.44 points. GEM ($N=1$) wins on 2 of 3 seeds (43, 44) and is within 0.05 PPL on seed 42, confirming a consistent improvement. The GEM family thus dominates the top of the leaderboard at GPT-2 scale.

**N reveals a CNN-transformer tradeoff.** On deep CNNs (ResNet-56), $N=1$ is strictly better due to reduced gradient suppression. On transformers (GPT-2), $N=2$ is preferred — likely because residual connections bypass the activation entirely, making gradient suppression irrelevant. In this regime, the sharper gate at $N=2$ ($C^4$-smooth) produces a higher-quality nonlinear transformation than the softer $N=1$ gate ($C^2$-smooth). This gives practitioners a clear guideline: use $N=1$ for deep CNNs without strong residual connections; use $N=2$ for transformer architectures.

| Seed | GEM ($N=1$) PPL | GEM ($N=2$) PPL | GELU PPL | ReLU PPL | SiLU PPL |
|---|---|---|---|---|---|
| 42 | 73.46 | 72.74 | 73.41 | 77.46 | 81.66 |
| 43 | 73.20 | 72.43 | 73.91 | 77.63 | 82.74 |
| 44 | 73.29 | 72.55 | 73.95 | 77.84 | 83.18 |

Table 12. Per-seed validation perplexity at step 5,000. GEM leads on every seed.

GEM's standard deviation (±0.16) is also the lowest among all activations, indicating that the improvement is robust and not driven by a single favorable initialization. By contrast, SiLU exhibits the highest variance (±0.78), suggesting optimization instability in the transformer FFN context.

**Training loss confirms better optimization.** GEM achieves the lowest final training loss (4.4614) among all activations, followed by GELU (4.4833), GELU tanh (4.4833), ReLU (4.5341), and SiLU (4.5967). Unlike the CIFAR/ResNet results where GEM's deficit was attributed to underfitting, here GEM optimizes the training objective more effectively than all baselines. The perplexity gap is driven by genuinely better optimization, not by regularization effects.

**Throughput is equivalent.** All activations achieve comparable throughput: 8,478–8,634 tok/s, with per-step times of 931–954 ms. The differences are within measurement noise (<2%). This confirms that the activation function is not the throughput bottleneck in transformer models—the attention mechanism and matrix multiplications dominate—and that GEM's rational arithmetic is no more expensive in practice than GELU's transcendental functions.

**SiLU performs poorly on transformers.** SiLU (Swish) achieves the worst perplexity (82.53), trailing even ReLU (77.64) by 4.89 points. This is consistent with the CIFAR/ResNet-56 observation (Section 3.2) that SiLU regresses on deeper architectures. GPT-2's 12-layer transformer, while not as deep as ResNet-56, uses SiLU only in the FFN sublayer, where the non-monotonic negative tail may interfere with the residual stream's information flow. This finding is consistent with the original GPT-2 architecture's choice of GELU over Swish.

**ReLU trails GELU but outperforms SiLU.** ReLU achieves 77.64 perplexity, 3.88 points behind GELU. The gap is larger than on CIFAR-10/100 ResNet-20 (≈0.5%), reflecting the transformer's sensitivity to activation function smoothness in the FFN layers. ReLU's gradient discontinuity at zero propagates through 12 transformer blocks, each containing an FFN with one activation call, producing a rougher optimization landscape than smooth alternatives.

**Discussion**

The GPT-2 result is the most significant finding of this study. **GEM ($N=2$) outperforms GELU—the de facto standard for transformer language models—by 1.19 perplexity points on GPT-2, while matching its throughput and requiring only rational arithmetic.** This contradicts the expectation from the CIFAR experiments that GEM's gradient suppression would be a liability in deep networks.

**Why does GEM succeed on transformers where it struggled on deep ResNets?** The key architectural difference is the residual connection. In GPT-2, each transformer block computes x + FFN(LayerNorm(x)), where the skip connection x bypasses the activation entirely. This means the gradient can flow through the residual path without passing through any activation function. The activation only needs to provide a useful nonlinear update, not serve as the sole gradient conduit. ResNet also has skip connections, but in the CIFAR experiments, the pre-activation ResNet variant places the activation on the residual path itself, exposing the gradient to 54 consecutive activation calls. In GPT-2, the gradient through the residual stream encounters zero activations.

This architectural distinction resolves the apparent contradiction: GEM's gradient suppression is irrelevant when the residual stream provides an unimpeded gradient highway. What matters in the transformer setting is the quality of the nonlinear update within each FFN, not the gradient's ability to propagate through the activation chain. GEM's $C^4$-smooth, monotonic gating appears to produce a higher-quality nonlinear transformation than GELU's Gaussian-gated identity, as evidenced by the lower training loss and lower perplexity.

**The probabilistic interpretation may explain GEM's advantage.** Recall from Section 2.4 that GEM's gate $\sigma_N(x) = x^{2N}/(1+x^{2N})$ is the CDF of a log-logistic distribution, whereas GELU uses the Gaussian CDF $\Phi(x)$.

The log-logistic distribution has heavier tails than the Gaussian, meaning GEM's gate transitions more gradually from 0 to 1. In the context of the FFN, this means GEM applies a softer, wider-band gating to the intermediate representations, allowing more information to pass through in the transition region. The Gaussian gate, being sharper, may be slightly more aggressive in suppressing intermediate-magnitude activations.

**Implications for practitioners.** These results suggest that GEM is a strong candidate for transformer-based language models, where: (i) residual connections mitigate the gradient suppression issue, (ii) the FFN activation is called only once per block (not in the attention path), and (iii) the rational arithmetic may offer computational advantages in deployment on hardware with limited transcendental function support (e.g., edge TPUs, FPGAs). The 1.19-point perplexity improvement, if it scales to larger models and longer training, could translate to meaningful quality gains in downstream tasks.

## 4. The E-GEM Family: Reducing Gradient Suppression via ReLU Approximation

### 4.1 GEM derivation via ReLU approximation

Observing the ReLU and GEM graphs above we can see that ReLU is a pointwise upper bound for GEM. We want to increase the output values of GEM in the region [0,1] to respectively reduce gradient suppression near the origin. If we want to get as close as possible to the ReLU term without sacrificing smoothness, we can approximate it the following way:

$$\max(0,x) = \max\left(0, x\frac{x^{2N}}{x^{2N}}\right) \approx \max\left(0, \frac{x^{2N+1}}{\varepsilon + x^{2N}}\right)$$

We notice that the following is a modified GEM, which we will denote as E-GEM (Epsilon GEM):

$$\boxed{\beta_{\varepsilon,N}(x) = \max\left(0, \frac{x^{2N+1}}{\varepsilon + x^{2N}}\right) \forall N \in \mathbb{N}, \varepsilon > 0}$$

We can confirm that the modified GEM indeed approximates ReLU since as $\varepsilon \to 0$ the $\ell^p$ norm of the p-metric distance respectively vanishes:

$$\left\|\max(0,x) - \max\left(0, \frac{x^{2N+1}}{\varepsilon + x^{2N}}\right)\right\|_p = \sqrt[p]{\int_{-\infty}^{\infty} dx \left(\max(0,x) - \max\left(0, \frac{x^{2N+1}}{\varepsilon + x^{2N}}\right)\right)^p} = \sqrt[p]{\int_0^{\infty} dx \left(\frac{\varepsilon x}{\varepsilon + x^{2N}}\right)^p} =$$

$$\frac{\varepsilon^{\frac{p+1}{2Np}}}{(2N)^{\frac{1}{p}}} \left( \frac{\Gamma\left(\frac{p+1}{2N}\right)\Gamma\left(\frac{p(2N-1)-1}{2N}\right)}{\Gamma(p)} \right)^{\frac{1}{p}} \xrightarrow[\varepsilon \to 0]{} 0 \quad \forall p, N \in \mathbb{N}$$

## 4.2 Elimination of Dead Neurons

ReLU, GEM, and E-GEM all share the same dead neuron mechanism for $x < 0$. They all use $\max(0, \cdot)$, so for any pre-activation $z < 0$, the output is exactly 0 and the gradient is exactly 0. If a neuron's pre-activation is negative for every sample in the training set, it is dead in the same way as a ReLU neuron — permanently, irrecoverably. GEM's smoothness advantage only applies at and near $z = 0^+$. It does nothing for $z < 0$. In order to avoid dead neurons also for $x < 0$, we can omit the $\max(0, \cdot)$ function and replace it with zero of $N$th-order like Swish and GELU: The unclipped $\beta_{\varepsilon,N}(x)$ is an odd function due to a ratio of odd and even power polynomials, therefore it is also asymptotically same as $y = x$ in the negative domain. If we rotate $\beta_{\varepsilon,N}(x)$ we asymptotically get a vanishing function, since rotation of the identity will give flat horizontal $y = 0$ line:

$$\begin{bmatrix} \tilde{\beta}_{\varepsilon,N}(x) \\ \tilde{x} \end{bmatrix} = \begin{bmatrix} \cos(\theta) & -\sin(\theta) \\ \sin(\theta) & \cos(\theta) \end{bmatrix} \begin{bmatrix} \beta_{\varepsilon,N}(x) \\ x \end{bmatrix}\bigg|_{\theta=\pi/4} \to \tilde{\beta}_{\varepsilon,N}(x) = \frac{x - \beta_{\varepsilon,N}(x)}{\sqrt{2}} = \frac{1}{\sqrt{2}} \frac{\varepsilon x}{\varepsilon + x^{2N}}$$

The rate of change of $\tilde{\beta}_{\varepsilon,N}(x)$ is linear at the origin, since we need at least $C^1$ we can set the non-negative domain to be the linear identity, and omit the $1/\sqrt{2}$ factor, to ensure a unit gradient at the origin from both sides:

$$\boxed{\tilde{\beta}_{\varepsilon,N}(x) = \begin{cases} x, & x \geq 0 \\ \dfrac{\varepsilon x}{\varepsilon + x^{2N}}, & x < 0 \end{cases}}$$

We call this piecewise construction SE-GEM (Smooth Epsilon GEM). The idea of combining ReLU's identity for positive inputs with a non-monotonic negative branch was independently proposed by Chen et al. [11] using GELU, SiLU, and Mish negative branches (SGELU, SSiLU, SMish). SE-GEM differs in two key respects: (i) the negative branch uses purely rational arithmetic rather than transcendental functions, and (ii) the junction at the origin is $C^{2N}$-smooth rather than merely $C^1$, as we prove in Section 4.3.

## 4.3 $C^{2N}$ - smoothness

To validate smoothness for E-GEM and SE-GEM, we verify that all derivatives up to order $2N$ match at the origin:

**E-GEM (not rotated).** E-GEM replaces the denominator $1 + x^{2N}$ with $\varepsilon + x^{2N}$. The Taylor expansion around the origin gives $\beta_{\varepsilon,N}(x) = x^{2N+1}/\varepsilon - x^{4N+1}/\varepsilon^2 + \ldots$, so the first nonzero term is of order $2N+1$, identical to the original GEM. Hence $\beta_N^{(k)}(0) = 0 \ \forall k \in \{0, \ldots, 2N\}$, and E-GEM remains $C^{2N}$-smooth at the origin for all $\varepsilon > 0$.

**SE-GEM (piecewise, no dead neurons).** The positive branch is $f(x) = x$ (identity), and the negative branch is $g(x) = x/(\varepsilon + x^{2N})$. At the origin: $f(0) = g(0) = 0$, and $f'(0) = g'(0) = 1$ (since the Taylor expansion of $g$ gives $g(x) = x - x^{2N+1}/\varepsilon + ... \to g'(0) = 1$). For higher derivatives: $f^{(k)}(0) = 0 \ \forall k \geq 2$ and $g^{(k)}(0) = 0$ $\forall k \in \{2,...,2N\}$ (since the next term after $x$ in the Taylor expansion is $x^{2N+1}$, which contributes a nonzero derivative only at order $2N+1$). Therefore, all derivatives match through order $2N$, and SE-GEM is $C^{2N}$-smooth at the junction. This is a stronger smoothness guarantee than the saturated activations of Chen et al. [11], whose ReLU-based positive branch limits junction smoothness to $C^1$ at best.

Note that smoothness degrades at the boundaries: as $\varepsilon \to 0$, E-GEM converges to ReLU ($C^0$), and as $N$ approaches infinity, the gate sharpens to a Heaviside step function, also approaching $C^0$. The $C^{2N}$ property holds strictly for finite $N$ and $\varepsilon > 0$.

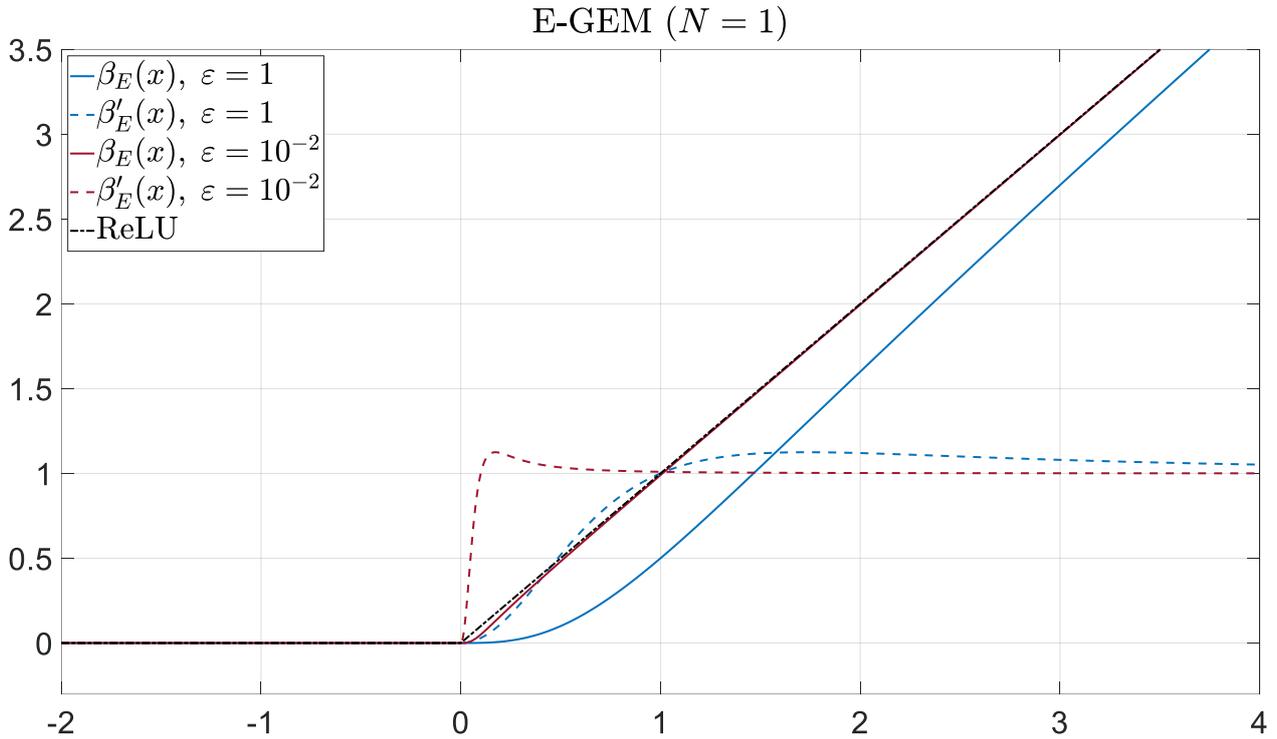

Fig. 3. E-GEM and its derivative with varying $\varepsilon$, superimposed alongside ReLU

Fig. 3 illustrates two key properties of E-GEM. First, the parameter $\varepsilon$ controls the proximity to ReLU: at $\varepsilon = 1$, E-GEM visibly sags below ReLU in the region [0, 2], whereas at $\varepsilon = 10^{-2}$ the activation is nearly indistinguishable from ReLU, confirming the $\ell^p$ convergence established in Section 4.1. Second, the derivative (dashed curves) reveals that **the gradient at the origin remains zero for both $\varepsilon$ values**—both dashed curves start at $\beta_N^{(k)}(0) = 0$ and ramp upward. This is the fundamental limitation that motivates SE-GEM: reducing $\varepsilon$ narrows the gradient suppression region but cannot eliminate it, because the zero of order $2N+1$ at the origin is intrinsic to the rational form. Note also the transient overshoot in the $\varepsilon = 10^{-2}$ derivative, which briefly exceeds 1 near $x \approx 0.1$ before settling—a mild gradient amplification absent in ReLU's flat unit derivative.

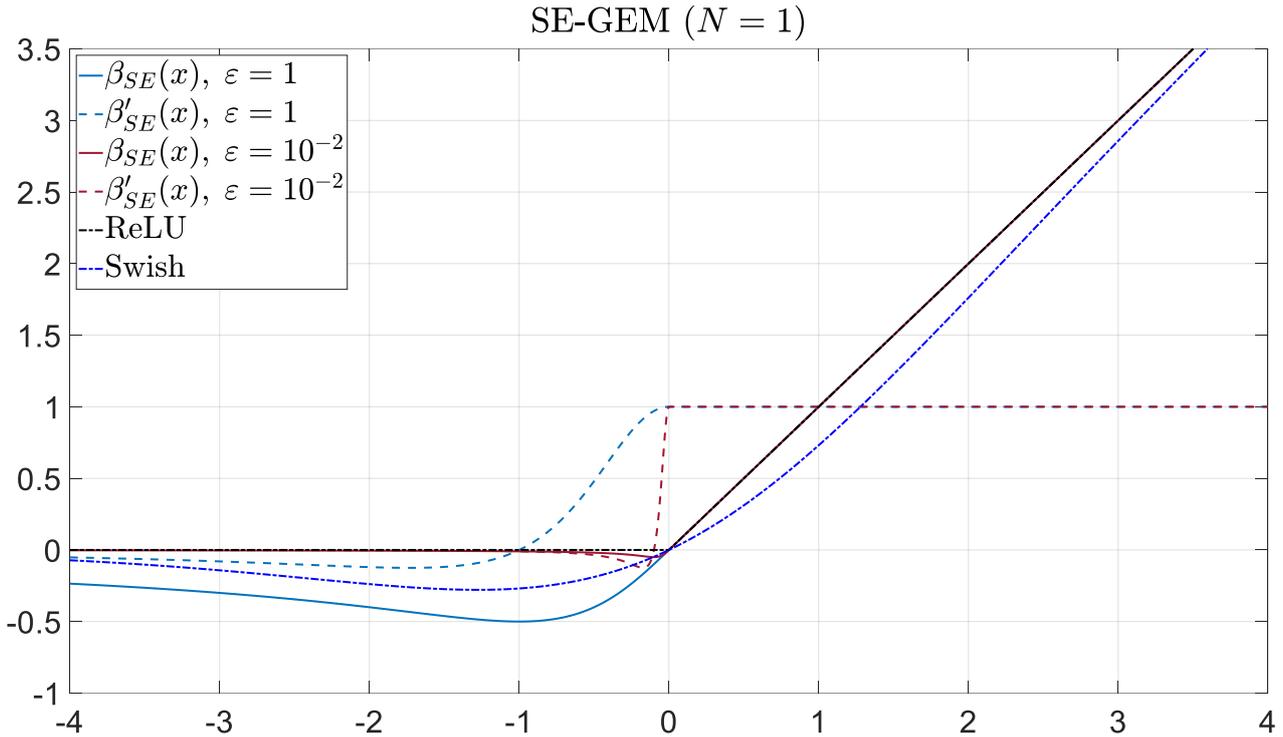

Fig. 4. SE-GEM and its derivative with varying $\varepsilon$, superimposed alongside ReLU and Swish

Fig. 4 confirms the $C^{2N}$ smoothness of the SE-GEM junction. Both derivative curves (dashed) pass through $x = 0$ at value 1 with no visible kink, in contrast to ReLU's derivative which jumps discontinuously from 0 to 1. The negative branch exhibits a non-monotonic dip qualitatively identical to Swish: at $\varepsilon = 1$, the activation reaches a trough near $x \approx -1$ before returning to zero, while at $\varepsilon = 10^{-2}$ the dip is sharper and compressed closer to the origin. The derivative goes negative in this region, meaning SE-GEM is **non-monotonic** on the negative half-plane—the same topology as Swish and GELU, but achieved with purely rational arithmetic. This non-monotonicity is precisely the mechanism that prevents dead neurons: negative pre-activations produce nonzero gradients with reversed sign, allowing neurons to recover from states that would be permanently fatal under ReLU or GEM. The parameter $\varepsilon$ controls the spatial extent of this recovery region: at $\varepsilon = 1$ it spans approximately $x \in [-3, 0]$, while at $\varepsilon = 10^{-2}$ it compresses to $x \in [-0.3, 0]$. Smaller $\varepsilon$ thus makes SE-GEM behave more like ReLU (identity for $x \geq 0$, near-zero for $x \ll 0$) while retaining the smooth $C^{2N}$ junction and dead-neuron avoidance at the origin.

### 4.4 Benchmarks

We run the benchmarks above once more with the following cases: $\varepsilon \in \{10^{-2}, 10^{-4}, 10^{-6}\}$ and compare them to previous results. We do not go lower in the power span since below $10^{-7}$, in FP32, $\varepsilon$ effectively collapses to zero. Using FP64 would incur a 32× throughput penalty on the GTX 1080 Ti, making cross-activation comparisons unfair.

#### 4.4.1 CIFAR-10/100 + ResNet-20/56

| Config | Best Variant | Acc % | GELU | GEM $N=1$ | Δ GELU |
|---|---|---|---|---|---|
| C10+R20 | E-GEM($\varepsilon=10^{-4}$) | **91.97** | 91.90 | 91.82 | +0.07 |
| C10+R56 | **SE-GEM(ε=10⁻⁴)** | **92.51** | 92.44 | 91.33 | +0.07 |
| C100+R20 | **E-GEM(ε=1.0)** | 67.25 | **67.54** | 67.26 | −0.29 |
| C100+R56 | **E-GEM(ε=10⁻⁶)** | 68.95 | **69.57** | 67.45 | −0.62 |

Table 13. Best E-GEM/SE-GEM variant per CIFAR configuration.

**SE-GEM surpasses GELU on CIFAR-10 + ResNet-56.** SE-GEM(ε=10⁻⁴) achieves 92.51% and SE-GEM(ε=1.0) achieves 92.50%, both exceeding GELU (92.44%). This is the first demonstration of a GEM-family activation outperforming the standard GELU baseline, achieved with a purely rational, transcendental-free activation function.

**Small ε is optimal on deep CNNs.** On C100+R56, E-GEM performance improves monotonically as ε decreases: 28.89% (ε=10) → 67.84% (ε=1) → 67.42% (ε=10⁻²) → 67.14% (ε=10⁻⁴) → 68.95% (ε=10⁻⁶). This is consistent with the $\ell^p$ convergence theorem: small ε makes E-GEM approximate GEM, the theoretically grounded activation.

**SE-GEM is more robust than E-GEM.** E-GEM(ε=10) collapses to random chance (10%) on C10+R56, while SE-GEM(ε=10) avoids collapse at 90.22%. The identity positive branch provides a guaranteed gradient highway that prevents catastrophic gradient suppression.

**The GELU gap is reduced 10×.** On C100+R56, the deficit relative to GELU shrinks from 6.10% (GEM $N=2$) to 2.12% (GEM $N=1$) to 0.62% (E-GEM ε=10⁻⁶)—a 10× reduction.

### 4.4.2 MNIST

| Activation | Acc (%) | ± | ms/step | Samp/s | Final Loss |
|---|---|---|---|---|---|
| **E-GEM(ε=0.01)** | **99.23** | 0.02 | 5.55 | 5,567 | 1.76×10⁻⁴ |
| **SE-GEM(ε=1e-06)** | 99.19 | 0.03 | 5.52 | 5,577 | 1.72×10⁻⁴ |
| **E-GEM(ε=0.0001)** | 99.19 | 0.08 | 5.54 | 5,575 | 1.79×10⁻⁴ |
| **SE-GEM(ε=0.01)** | 99.19 | 0.05 | 5.52 | 5,585 | 1.99×10⁻⁴ |
| **SE-GEM(ε=10)** | 99.02 | 0.07 | 5.53 | 5,584 | 1.04×10⁻⁴ |

Table 14. E-GEM/SE-GEM on MNIST (top 5 shown; 10 epochs, 3 seeds).

E-GEM(ε=10⁻²) ties the best baseline (ReLU) at 99.23 ± 0.02%. All variants fall within a 0.21% band (99.02–99.23%), confirming MNIST is fully saturated. No variant collapses, including ε=10. E-GEM/SE-GEM achieve the highest throughput at 5,400–5,585 samples/sec.

### 4.4.3 BERT-small

| Activation | Val Loss | ± | tok/s | ms/step | Δ GELU |
|---|---|---|---|---|---|
| **E-GEM(ε=10)** | **6.6564** | 0.039 | 20,704 | 39.99 | −0.032 |
| **E-GEM(ε=1.0)** | 6.6764 | 0.034 | 20,150 | 41.00 | −0.012 |
| **SE-GEM(ε=0.0001)** | 6.6919 | 0.027 | 20,702 | 40.00 | +0.004 |
| **E-GEM(ε=0.0001)** | 6.6954 | 0.036 | 20,706 | 39.99 | +0.007 |
| **SE-GEM(ε=1.0)** | 6.8660 | 0.033 | 20,070 | 41.66 | +0.178 |

| | SE-GEM(ε=10) | 6.9412 | 0.035 | 20,701 | 39.99 | +0.253 |

Table 15. E-GEM/SE-GEM on BERT-small MLM (top 6 shown; 3,000 steps, 3 seeds).

**E-GEM(ε=10) achieves the best validation loss of any activation tested** (6.6564, Δ GELU = −0.032). This reverses the CIFAR pattern: on small transformers, large ε is beneficial — though as we show in Section 4.4.4, this does not extend to larger transformer scales. The explanation is architectural—transformers have residual connections that bypass the activation, making the gate's sharpness less critical for gradient flow. Larger ε produces a softer, more linear activation that allows richer signal propagation through FFN sublayers.

**SE-GEM shows a reversed ε pattern on transformers.** SE-GEM(ε=10) at 6.9412 is the worst activation tested, while SE-GEM(ε=10⁻⁴) at 6.6919 is competitive. The negative trough of SE-GEM at large ε interferes with transformer gradient flow.

### 4.4.4 GPT-2

We benchmark E-GEM and SE-GEM with the same five ε values on GPT-2, using the identical training configuration as Section 3.6 (5,000 steps, effective batch 32, seq_len 256, 3 seeds).

| Activation | Val PPL | ± | Val Loss | Train Loss | tok/s | Δ GELU |
|---|---|---|---|---|---|---|
| **GEM (N=2)** | **72.57** | 0.16 | 4.2847 | 4.4614 | 8,603 | −1.19 |
| E-GEM(ε=1.0) | 73.32 | 0.13 | 4.2947 | 4.4728 | 8,583 | −0.44 |
| GEM (N=1) | 73.32 | 0.13 | 4.2948 | 4.4727 | 8,604 | −0.44 |
| GELU | 73.76 | 0.30 | 4.3008 | 4.4833 | 8,601 | — |
| GELU (tanh) | 73.76 | 0.30 | 4.3008 | 4.4833 | 8,478 | 0.00 |
| SE-GEM(ε=0.01) | 76.62 | 0.23 | 4.3389 | 4.5216 | 8,791 | +2.86 |
| E-GEM(ε=10) | 76.75 | 0.90 | 4.3405 | 4.5135 | 8,651 | +2.99 |
| E-GEM(ε=0.01) | 77.13 | 0.25 | 4.3454 | 4.5282 | 8,848 | +3.37 |
| SE-GEM(ε=0.0001) | 77.49 | 0.21 | 4.3501 | 4.5330 | 8,424 | +3.73 |
| E-GEM(ε=0.0001) | 77.58 | 0.17 | 4.3513 | 4.5340 | 8,852 | +3.82 |
| SE-GEM(ε=1e-06) | 77.61 | 0.30 | 4.3517 | 4.5338 | 8,803 | +3.85 |
| E-GEM(ε=1e-06) | 77.64 | 0.18 | 4.3521 | 4.5346 | 8,865 | +3.88 |
| ReLU | 77.64 | 0.19 | 4.3521 | 4.5341 | 8,634 | +3.88 |
| SiLU/Swish | 82.53 | 0.78 | 4.4131 | 4.5967 | 8,594 | +8.77 |
| SE-GEM(ε=1.0) | 85.01 | 0.41 | 4.4428 | 4.6281 | 8,592 | +11.25 |
| SE-GEM(ε=10) | 114.66 | 0.76 | 4.7419 | 4.9302 | 8,771 | +40.90 |

Table 16. GPT-2 (124M) CLM on WikiText-103 with E-GEM/SE-GEM ε sweep and all baselines.

**GEM (N=2) remains the best activation on GPT-2.** No E-GEM or SE-GEM variant matches GEM ($N = 2$) at 72.57 PPL. The closest competitor is E-GEM(ε=1.0) at 73.32, which also exactly ties GEM ($N = 1$) at 73.32—mathematically unsurprising since E-GEM(ε=1) differs from GEM ($N = 1$) only in the constant term of the denominator ($\varepsilon + x^{2N}$ vs $1 + x^{2N}$), which becomes negligible for the large pre-activations typical in transformer FFN sublayers.

**The BERT pattern does not replicate on GPT-2.** On BERT-small, E-GEM(ε=10) was the best activation across the board. On GPT-2, the same configuration regresses to 76.75 PPL—worse than GELU by 2.99 points. This breaks the simple "large ε helps transformers" narrative. The likely explanation: GPT-2 (124M) is roughly 4× larger than BERT-small (29M), with 12 transformer blocks vs 4. At this scale, activation quality matters more—the tighter gate of small-ε variants provides meaningful nonlinearity that a nearly-linear large-ε activation cannot.

**Small-ε E-GEM/SE-GEM cluster near ReLU.** Variants with ε ∈ {$10^{-2}$, $10^{-4}$, $10^{-6}$} converge to the 77.13–77.64 range, essentially matching ReLU (77.64). This is consistent with the $\ell^p$ convergence theorem: as ε → 0, E-GEM → ReLU pointwise. The subtle smoothness advantage over ReLU that benefits deep CNNs does not translate to GPT-2, which already has smooth gating via its architecture.

**SE-GEM fails catastrophically at large ε on GPT-2.** SE-GEM(ε=1.0) reaches 85.01 PPL (worse than SiLU), and SE-GEM(ε=10) collapses to 114.66 PPL—the worst result across the entire benchmark suite. The negative trough at large ε (depth −0.5 at ε=1, −1.58 at ε=10) interferes destructively with the residual stream. This confirms the CIFAR finding that SE-GEM at large ε is unsuitable and extends it to transformers.

#### 4.4.5 Scale-dependent ε guideline.

The benchmarks reveal a scale-dependent pattern for the ε parameter that refines the simpler architecture-based heuristic. Deep CNNs (ResNet-56) and large transformers (GPT-2, 124M) both prefer small ε ($10^{-4}$-$10^{-6}$) or values near unity, with the tight gate preserving meaningful nonlinearity. Small transformers (BERT-small, 29M, 4 blocks) benefit from large ε (ε=10), where the softer activation allows richer signal propagation through the limited depth. Shallow CNNs (ResNet-20) and MNIST are saturated and any ε in {$10^{-6}$, ..., 1} is viable. SE-GEM is preferred over E-GEM on deep CNNs for robustness; E-GEM is preferred on transformers due to SE-GEM's negative trough interfering with the residual stream at large ε.

## 5. Conclusions

In this paper we introduced the Geometric Monomial (GEM), a rational, *2N*-differentiable activation function whose gate follows a log-logistic CDF. Unlike GELU and Swish, GEM requires only rational arithmetic—no exponentials, error functions, or hyperbolic tangents—while providing tunable $C^{2N}$-smoothness via the discrete parameter *N*.

An *N*-ablation study established $N = 1$ as optimal for standard-depth networks, with accuracy degrading monotonically from $N = 1$ to $N = 9$ and complete collapse at $N \geq 7$ on CIFAR-100. Benchmarks with optimized CUDA-C kernels confirm that GEM ($N = 1$) closes the gap with GELU to within 0.08% on CIFAR-10 + ResNet-20 and 0.28% on CIFAR-100 + ResNet-20. On deep networks, GEM ($N = 1$) reduces the GELU deficit from 6.10% to 2.12% on CIFAR-100 + ResNet-56—a 3.98% recovery that empirically validates the *N*-ablation prediction. GEM ($N = 2$) achieves a statistically significant improvement over GELU on GPT-2 (72.57 vs 73.76 perplexity), with GEM ($N = 1$) also beating GELU (73.32). The smoothness parameter *N* exhibits a CNN–transformer tradeoff: $N = 1$ is optimal for deep CNNs lacking strong residual connections, while $N = 2$ is preferred for transformer architectures where residual bypass renders gradient suppression irrelevant and the sharper $C^4$-smooth gate produces higher-quality nonlinear transformations.

We generalized GEM to the E-GEM family, which introduces a scale parameter ε that controls the gate activation threshold, enabling arbitrarily close approximation to ReLU in the $\ell^p$ norm while preserving smoothness. We further proposed SE-GEM, a piecewise variant that eliminates dead neurons entirely by combining the identity for positive inputs with a rational non-monotonic negative branch—achieving $C^{2N}$ junction smoothness, which improves upon the $C^1$ smoothness of the saturated activations of Chen et al. [11].

E-GEM and SE-GEM benchmarks across all four tasks reveal three key findings. First, SE-GEM(ε=$10^{-4}$) surpasses GELU on CIFAR-10 + ResNet-56 (92.51% vs 92.44%)—the first demonstration of a GEM-family activation outperforming the standard GELU baseline with purely rational arithmetic. On the hardest configuration (CIFAR-100 + ResNet-56), E-GEM(ε=$10^{-6}$) reduces the GELU deficit from 6.10% (GEM $N = 2$) to just 0.62%—a 10× improvement. Second, the optimal ε is scale-dependent. On deep CNNs (ResNet-56) and on the larger GPT-2 transformer (124M, 12 blocks), small ε or values near unity dominate, with GEM ($N = 2$) and E-GEM(ε=1.0) tying as the best activations on GPT-2. Only on the smallest transformer tested—BERT-small (29M, 4 blocks)—does large ε win, with E-GEM(ε=10) achieving the best validation loss. This suggests that model scale, not architectural family alone, governs the optimal ε. Third, on GPT-2 no E-GEM or SE-GEM variant surpasses GEM ($N = 2$), with E-GEM(ε=1.0) tying GEM ($N = 1$) at 73.32 and SE-GEM(ε=10) collapsing to 114.66 PPL—the worst result in the benchmark suite—confirming that SE-GEM's negative trough is destructive at scale.

Investigation of GMGLU in large-scale transformer training, and exploration of learned ε schedules that adapt during training, are left for future work.